\documentclass[10pt,twocolumn,letterpaper]{article}

\usepackage{iccv}
\usepackage{times}
\usepackage{epsfig}
\usepackage{graphicx}
\usepackage{amsmath}
\usepackage{amssymb}
\usepackage[square,sort,comma,numbers]{natbib}

\usepackage[breaklinks=true,bookmarks=false]{hyperref}

\iccvfinalcopy 

\def\iccvPaperID{34} 

\ificcvfinal\fi

\begin{document}

\title{Indoor Depth Completion with Boundary Consistency and Self-Attention}

\author{Yu-Kai Huang\thanks{equal contribution} \qquad Tsung-Han Wu\footnotemark[1] \qquad Yueh-Cheng Liu \qquad Winston H. Hsu \\
National Taiwan University\\
{\tt\small \{r08922053, b05902013, r07922183, whsu\}@ntu.edu.tw}
}


\maketitle
\ificcvfinal\thispagestyle{empty}\fi

\begin{abstract}
Depth estimation features are helpful for 3D recognition. Commodity-grade depth cameras are able to capture depth and color image in real-time. However, glossy, transparent or distant surface cannot be scanned properly by the sensor. As a result, enhancement and restoration from sensing depth is an important task. Depth completion aims at filling the holes that sensors fail to detect, which is still a complex task for machine to learn. Traditional hand-tuned methods have reached their limits, while neural network based methods tend to copy and interpolate the output from surrounding depth values. This leads to blurred boundaries, and structures of the depth map are lost.
   Consequently, our main work is to design an end-to-end network improving completion depth maps while maintaining edge clarity. We utilize self-attention mechanism, previously used in image inpainting fields, to extract more useful information in each layer of convolution so that the complete depth map is enhanced. In addition, we propose boundary consistency concept to enhance the depth map quality and structure. Experimental results validate the effectiveness of our self-attention and boundary consistency schema, which outperforms previous state-of-the-art depth completion work on Matterport3D dataset. Our code is publicly available at  \href{https://github.com/tsunghan-wu/Depth-Completion}{\color{blue}{https://github.com/tsunghan-wu/Depth-Completion}}
\end{abstract}

\section{Introduction}
In this task, we take single input RGB and raw sensing depth to complete depth value of missing parts. Depth sensing is widely used in real-world applications from autonomous driving, robotics, augmented reality to scene reconstruction. Most of the applications are for recognition purpose. However, commercially available RGB-D cameras, such as Microsoft Kinect and Intel RealSense, still fail on sensing depth map without corruption or noise along object edges. Sensors produce large amounts of missing pixels when surface is shiny, transparent, too close or too far. In indoor dataset Matterport3D \cite{chang2017matterport3d}, for example, over $15\%$ of values in raw depth are missing. As a consequence, completing large missing parts of sensor depth is of crutial importance.

Traditional works reconstruct depth value by combining multiple views of sensor data. Simultaneous localization
and mapping (SLAM) \cite{durrant2006simultaneous, bailey2006simultaneous} method incrementally builds a consistent depth map of the environment while simultaneously determines location within this map. Previous methods \cite{henry2012rgb, engel2014lsd} utilizes SLAM to estimate more accurate depth map. In our work, we do not consider multiple views of reconstruction. Rather, we put emphasis on single view for depth prediction, because single view prediction is more efficient when consider inference time and computing power.

\begin{figure}[t!]
    \centering
    \includegraphics[width=\linewidth]{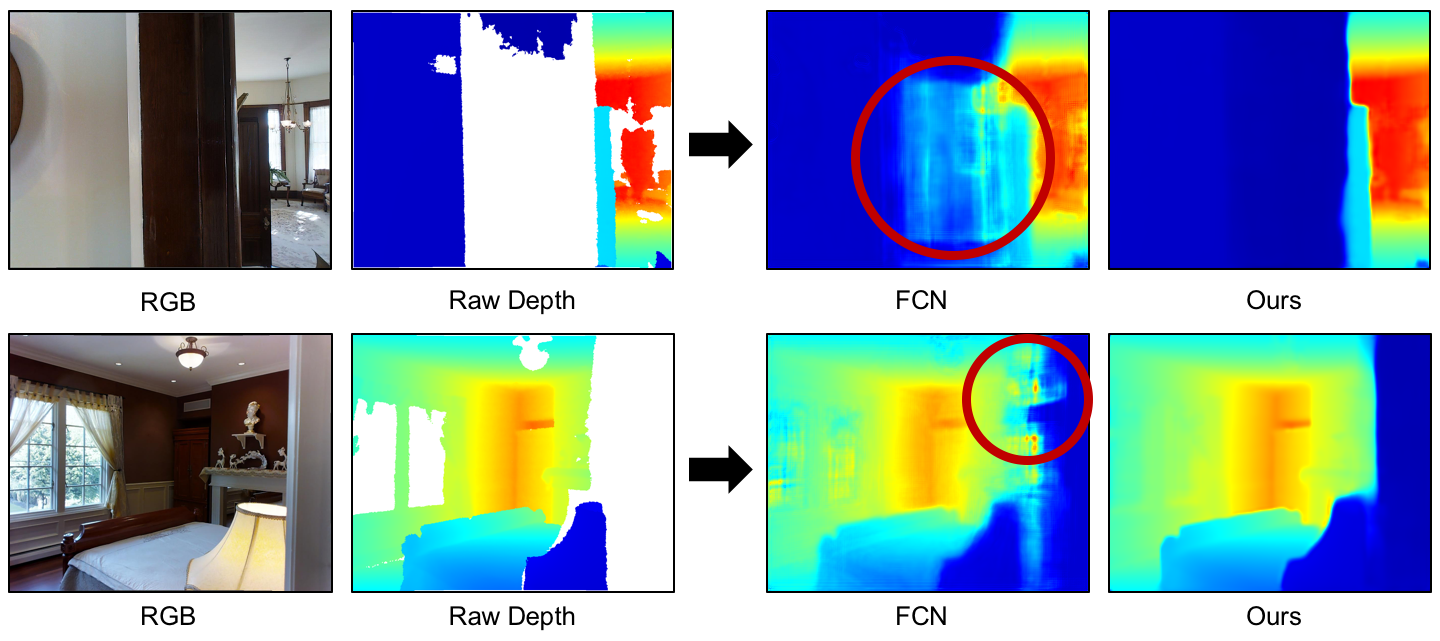}
    \caption{Our work produces clearer structures and overcomes the hardness that previous works suffer from: interpolation and blurred boundaries. On the first row, our model provides clear complete depth but FCN interpolates depth value (marked in red) at the large missing area of raw depth map. On the second row, our model can assure boundary sharpness, while FCN generates blurred boundaries.}
    \label{fig:figure1}
\end{figure}

For single view of depth completion, previous works that tried to solve the problem have their own issue. Traditional mathematical methods \cite{matsuo2015depth, doria2012filling} contain much hand-tuning hyper-parameters and reach their limits in filling high quality depth holes. Recent deep learning methods \cite{eigen2014depth, xie2016deep3d} performed well yet only learned to interpolate or copy-and-paste depth value from neighboring pixels. Blurred boundaries and structure can be easily recognized by visualization of output depth. In Figure \ref{fig:figure1}, the visualization results show the above issue of copy-and-paste, interpolation, blurred boundaries and structure. We conclude the issues of deep learning based methods in two orientations: estimating precise depth value and producing clear structure.

In this paper, we solve the issues mentioned above. Previous works pass the whole feature maps into the network and perform convolution. However, to complete the missing depths, some regions may be more important and some may be minor details depending on the semantics and geometric meanings of one scene. Inspired by Yu \etal~\cite{yu2018free}, we leverage the self-attention network to encourage the model to pay attention on relevant parts, especially the semantics of the scene. Our model benefits from the attention mechanism and outputs depth maps with comprehensive consideration of the attentioned parts. Furthermore, with useful information from depth representation (surface normals and occlusion boundaries), proposed by \cite{zhang2018deep}, our self-attention model meets the needs of estimating more precise depth value rather than just interpolation.

In addition, we solve the issue of vague structures with boundary consistency concept. In order to generate clear depth structures, we equip another network to predict occlusion boundaries given the output depth from depth completion network. As a result, the depth completion network is indirectly refined to preserve clear boundaries in the output depth, making the complete depth image to be more structured and conformed to realistic situation. To the best of our knowledge, we are the first to apply self-attention mechanism and regard boundary as the main feature in indoor depth completion task.

On the whole, our main contributions of our work are:
\begin{itemize}
    \item Enhance depth completion task using self-attention mechanism, which has never been used in RGB-D estimation before.
    \item Propose a novel idea, boundary consistency, to produce depth map of clear structure. 
    \item Reach state-of-the-art performance on Matterport3D \cite{chang2017matterport3d} on RGB-D depth completion task.
\end{itemize}

Extensive ablation study and visualization results validate our proposed idea and concept. The promising results on depth completion task makes contribution to  recognition applications such as robotics and autonomous driving. Those applications suffer from various types of degradation of low-quality visual data, including large missing holes of sensor depth map. Our work enhances the robustness for recognition on sensor depth, and, consequently, contributes to real-world recognition from low-quality images.

\section{Related Work}

We introduce order as follows: depth estimation, super-resolution, reconstruction from sparse samples, image inpainting, and, finally, our task depth completion. 
\subsection{Depth Estimation}
Depth estimation predicts depth value from monocular RGB image. This is a long-established problem in computer vision history. However, with little information of a single view image, this is still a hard task even for human beings to answer the 
exact depth distance \cite{mingolla1986perception}.
Classic methods like Shape from Shading \cite{zhang1999shape, mingolla1989perception} and depth from focus \cite{grossmann1987depth, suwajanakorn2015depth} elaborated physical and mathematical property about light shading and focal setting at each pixel. Recent works extended classic methods by machine learning, like deep depth from focus \cite{hazirbas2018deep} and deep estimation based on fourier domain analysis \cite{lee2018single}. Fully connected convolution (FCN) networks are used to predict depth map \cite{laina2016deeper} or refine coarse-scale depth value \cite{eigen2014depth}. 

Previous works give us insight into the methods to solve the task, but different from what our topic focuses on. Specifically, depth estimation generate raw depth based on RGB image, while we utilize RGB to improve raw depth to become complete.

\subsection{Depth Super-Resolution}
The focus of depth super-resolution is the enhancement of spatial resolution. Generally, raw depth data are regularly sampled or quantized with low resolution. Yang \etal \cite{yang2007spatial} iteratively refined depth map in terms of both its spatial resolution and depth precision with bilateral filtering. Mac \etal \cite{mac2012patch} utilized Markov random field to select candidate of depth patch. 
Some other works are shape-from-shading \cite{han2013high, yu2013shading}, rigid body self-similarity \cite{hornacek2013depth}, and deep learning based network \cite{riegler2016atgv}. 

Different from depth completion, these works take  complete but low-resolution depth map as input. In our work, we aim at recovering the large missing parts of the depth map, which cannot be recovered by super-resolution methods.

\subsection{Depth Reconstruction from Sparse Samples}
Depth reconstruction from sparse samples is an interesting topic that tries to reconstruct full depth map from sparse one.
Restrictions of camera costs and power consumption make it a necessary research topic. The difficulties of the topic are how to maintain decent performance, while saving costs and energy. Ma's method \cite{mal2018sparse} can be used as a plug-in module in sparse SLAM and visual inertial odometry algorithms, creating more accurate, dense point cloud. Also, Ma \cite{ma2018self} proposed a self-supervised framework without the need for dense labels, achieving great performance on KITTI \cite{geiger2013vision} dataset. Some works combined semantic segmentation \cite{jaritz2018sparse} to improve the prediction. 

\begin{figure*}[t!]
    \centering
    \includegraphics[width=\textwidth]{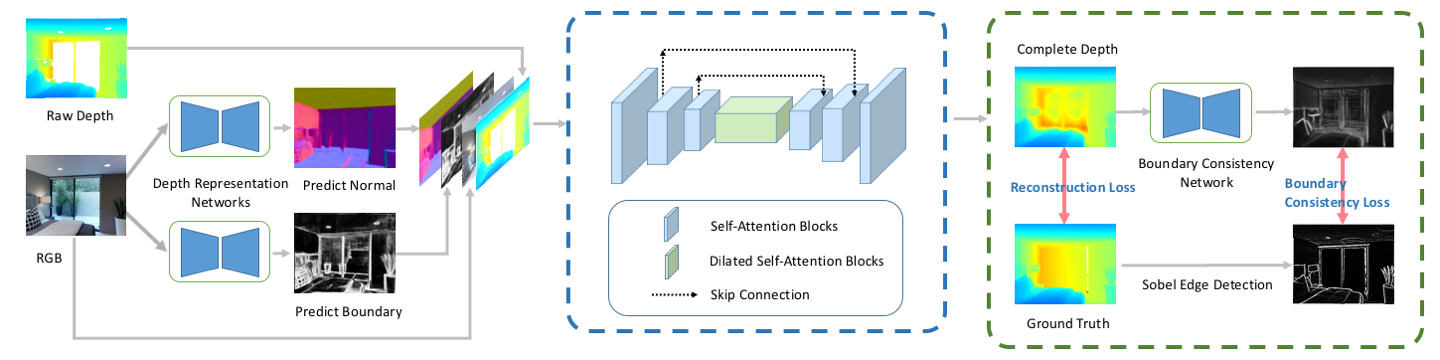}
    \caption{Overview of the network architecture. Main improvement of our method are twofold: self-attention mechanism \ref{sec:self-attention} (blue dashed box)  and boundary consistency \ref{sec:boundary_consistency} (green dashed box). Self-attention network pays attention on useful feature information. On the other hand, boundary consistency loss forces attention network to preserve boundary information, allowing network to reconstruct more structured depth map.}
    \label{fig:network}
\end{figure*}

\subsection{Image Inpainting}
Image inpainting is related to depth completion in some ways. For example, free-form image inpainting \cite{yu2018free} allows users to erase values of a RGB image, and then network inpaints the missing parts of RGB values as real as possible. Compared to our task, the same part is the missing holes of depth sensor are similar to the erased parts of RGB image. The different part is depth completion network is trained to learn precise value of depth map supervised with correct answer, while image inpainting results do not require a correct answer but a real and reasonable one. Self-attention mechanism, which applies attention on each layer of convolution, is useful and powerful for image inpainting task. Yu \etal utilized self-attention mechanism on partial convolution \cite{liu2018image} and gated convolution \cite{yu2018free} and made astonishing performance on the inpainting image. We introduce the concept of self-attention on our task.

\subsection{Depth Completion}
Depth completion is to fill in missing holes in the relatively dense
depth images. The definition of depth completion concept is proposed by Zhang \etal \cite{zhang2018deep}. Earlier works \cite{silberman2012indoor, camplani2012efficient} evaluated performance on pixels captured by commodity RGB-D cameras. Under the circumstances, they can at best output raw depth (captured by the sensor), which contains large missing holes. Zhang extended the term "Depth Completion" to predicting the complete depth map generated from multi-view reconstruction.


In Zhang's work, he proved surface normals and occlusion boundaries from RGB are two geometric representations best for deep depth completion. 
Zhang also claimed that deep regression method simply learns to copy and interpolate depth value, so he optimized loss with sparse Cholesky factorization scheme \cite{strang1986introduction} and reach state-of-the-art performance. 

In our work, we re-verify the copy and interpolation issue of standard FCN. We overcome the obstacles that FCN encounters, and our model can generate depth map with clear boundaries and structures through an end-to-end neural network. Our approach outperforms the previous methods. Compared to Zhang's, we have faster inference time by replacing the Cholesky optimization, which is more desirable for real world applications.


\section{Method}

In this paper, we investigate how to complete indoor depth image given a single RGB-D image with effective deep learning method. Our work mainly focuses on the following two questions: "How to overcome the difficulty that network simply learns to interpolate depth value?" and "how to make clear structures of depth image?"

\subsection{Self-Attention Mechanism} \label{sec:self-attention}
The first issue we address is that previous deep learning methods tend to copy and interpolate nearby depth value as outputs as shown in Figure \ref{fig:figure1}. Since in most cases, predicting average depth values for unobserved area makes the loss drop rapidly, so the network can easily fall into local minima of copy and interpolation instead of predicting precise depth values.
To solve this problem, we propose to use self-attention mechanism on each convolution layer. The network is allowed to focus on precise feature values at each convolution stage and forward useful information. With self-attention, the network can be more robust in predicting precise depth values and would not be easily trapped in local minima of copy and interpolation.

We use gated convolution, proposed by Yu \etal \cite{yu2018free}, as our backbone component. Yu \etal used gated convolution on free-form image inpainting task. We use gated convolution on our task, since, first, depth completion task aims to complete missing depth value, which is similar to free-form masks from users. Additionally, completing precise depth value is similar to inpainting with realistic RGB pixel values. As a result, we gain insight from the strong connection between the two different tasks. 

For an input $I$ of a convolution block, and convolution blocks for feature extraction $Conv_f$ and for gating $Conv_g$, the mathematical formulation for self-attention mechanism can be presented as:
\begin{align}
        \text{Gating} &= \sigma\left(SN(Conv_g(\text{I}))\right)\\
        \text{Feature} &= \phi \left(SN(Conv_f(\text{I}))\right)\\
        \text{Output} &= \text{Feature} \odot \text{Gating}
\end{align}
where $\sigma$ is sigmoid function, SN is spectral normalization, $\phi$ is any activation function, and $\odot$ is the pixel-wise multiplication. The self-attention mechanism is in the presence of an additional gating operation that Feature $\odot$ Gating. It allows the model to learn effective dynamic feature selection, and also highlights the feature meanings for each channel map and each spatial location. 

Since $Conv_g$ learns to identify region that is useful and important, according to the above equation, the model preserves useful region of Feature in the Output.
Therefore, the gated convolution can predict more accurate depth values by paying more attention on local and detailed information extracted by the self-attention mechanism.

\subsection{Boundary consistency} \label{sec:boundary_consistency}
Ordinary FCN methods are likely to generate depth image of copy and interpolation as explained above. This causes the output depth to have blurred boundaries and structures. For instance, line segments may be slightly distorted and object occlusion boundary is vague in most cases. Since depth completion is a complex regression problem, FCN models can hardly learn to generate depth map of sharp boundaries where neighboring pixel values differ greatly.
In order to fix the problem, we introduce boundary consistency to encourage models to learn clear boundaries and structures.

Boundary consistency constrains the output depth map to have boundary information by constructing another network, boundary consistency network. The boundary consistency network, modified from U-Net \cite{ronneberger2015u}, takes complete depth as input and predicts the occlusion boundary. It is trained end-to-end along with self-attention network. The boundary consistency network encourages the self-attention network to generate sharp edges so that boundary consistency network can estimate more precise boundary values. With moderate balance between the weights of losses, our self-attention network successfully completes depth map with sharper and clearer structures. The overview pipeline is shown in Figure \ref{fig:figure1}, we add an additional encode-decode boundary consistency network to the original self-attention network. We name it "boundary consistency" because we use occlusion boundary as depth input feature and require our model to preserve the boundary information in the complete depth map.


The boundary estimation result is supervised by Sobel \cite{kittler1983accuracy} boundary from ground truth depth image. There are two reasons we use Sobel as our target loss. (1) The generated edges are sharp, clear and few variate compared to some other methods. (2) Sobel Algorithm can extract occlusion boundaries without being too sensitive to the noise. 

With the help of occlusion boundary ground truth from Sobel, our generated depth map can learn sharper and clearer structures, preventing outputs like blurred boundaries and interpolation.

\subsection{Depth Representation}
In Zhang \etal \cite{zhang2018deep} work, he mentioned two important depth representations: surface normals and occlusion boundaries. Surface normals relate strongly to local lighting variations, providing more information about local surface properties. Occlusion boundaries also represent local texture features observed from RGB images. Both of the two representations can be robustly detected by networks, called depth representation network in the following. Because it is proven by \cite{zhang2018deep} that surface normals and occlusion boundaries are powerful, we simply combine the two representations with RGB and raw depth as the input of self-attention network.

\begin{figure*}[t!]
    \centering
    \includegraphics[width=\textwidth]{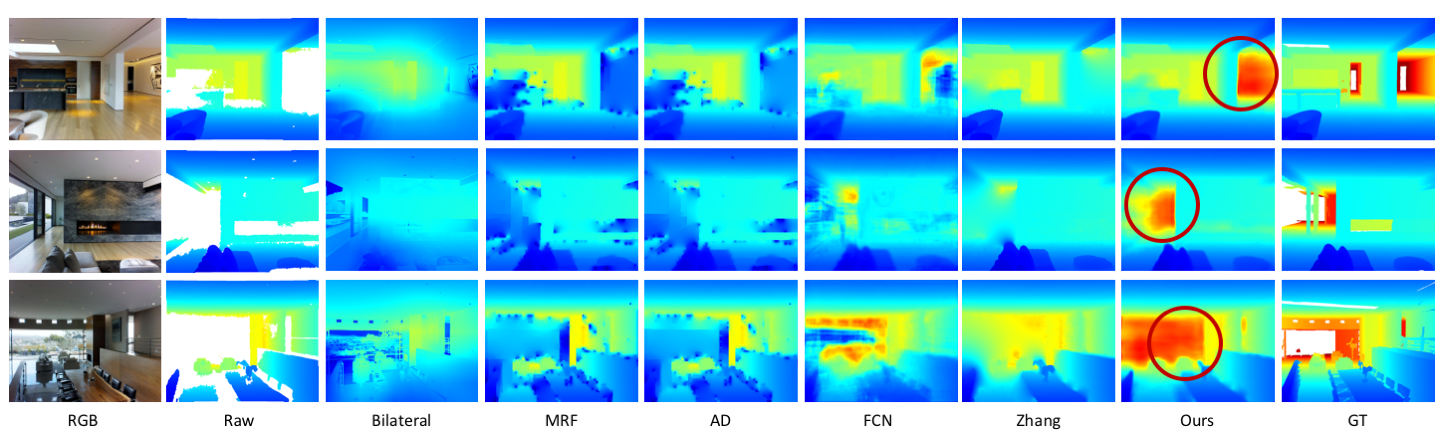}
    \caption{In the visualization result, we can see that our method learns better on the geometric meaning. For example, on the first row, our model notices the corridors on the right hand side next to the pillar and judge the depth to be deep (mark in red). Other methods tend to interpolate the large vacancy with nearby values, like Zhang fills in light blue depth value.}
    \label{fig:performance}
\end{figure*}

\subsection{Loss Functions} \label{sec:loss_function}
The following section would introduce each component of our loss in detail. The total loss can be written as:
\begin{equation}
    L = L_{SA} - \lambda_{S} L_{S} + \lambda_{BC} L_{BC} + \lambda_N L_{N} + \lambda_B L_{B} 
\end{equation}

Given input RGB, depth representation networks predict normals $N$ and boundaries $B$. The depth representation losses are $L_{N} = \frac{1}{|p|}\sum_{p} \|N(p) - N_0(p)\|^2$ and $L_{B} = \frac{1}{|p|}\sum_{p} \|B(p) - B_0(p)\|^2$, where $N_0$, $B_0$ are ground truths and $p$ is a pixel in image channel.

Then, RGB, depth representations and raw depth are combined as input of self-attention network which predicts complete depth $D$. The ground truth depth $D_0$ comes from rendering multiple raw depth views of the same scene. The loss for self-attention network is $L_{SA} = \frac{1}{|p \in obs|}\sum_{p \in obs} \|D(p) - D_0(p)\|$. Note that we only consider observed pixels, which have valid depth values in ground truth $D_0$, as mentioned in~\cite{zhang2018deep}. Missing parts in ground truth depth map are ignored. We denote $p \in obs$ to be observed pixels.

To enhance structural information, we add small fraction of structural related loss $L_S$, Structural Similarity Index (SSIM) \cite{wang2004image}. SSIM measures quality assessment based on the degradation of structural information. In our task, the higher SSIM index is, the more structured the complete depth is. With SSIM, we expect our network to learn higher quality and undistorted depth map with better structure.

Finally, complete depth $D$ is fed into boundary consistency network and outputs $B$. The ground truth depth $D_0$ generates ground truth $B_0$ with Sobel \cite{kittler1983accuracy}. The boundary consistency loss is $L_{BC} = \frac{1}{|p|}\sum_{p} \|B(p) - B_0(p)\|$.

\section{Experimental Results}

\subsection{Dataset}
We did not use the widely-known NYUv2 dataset \cite{silberman2012indoor}, since the dataset does not provide complete ground-truth depth map for a single captured RGB-D image. 
Instead, we ran the experiments on Matterport3D dataset \cite{chang2017matterport3d} to evaluate our proposed methods. Matterport3D is an indoor large-scale RGB-D dataset with 10.8k real panoramic views and 90 real indoor scenes. The dataset is scalable, diverse and representative of indoor depth completion task. We use the same training and testing lists as Zhang \cite{zhang2018deep} did. We remove large bias image pairs (about $5\%$), and, finally, the dataset contains about 1M training data and 474 testing data. The ground truth of Matterport3D is generated from Zhang by multi-view reconstruction; that is, multiple camera views of the same scene can reconstruct a more complete indoor information, including depth map.

\subsection{Evaluation Metrics}
We use similar evaluation metrics in \cite{zhang2018deep}. As mentioned in \ref{sec:loss_function}, we calculate loss based on observed data; that is, observed pixel values in ground truth depth. Given ground truth depth $D_0$ and complete depth $D$, the metrics include: 
\begin{itemize}
    \item Root Mean Square Error (RMSE):
\begin{equation}
    \sqrt{\frac{1}{|obs|}\sum_{p\in obs}\|D(p) - D_0(p)\|^2}
\end{equation} 

\item Mean Error (Mean): 
\begin{equation}
    \frac{1}{|obs|}\sum_{p\in obs}\|D(p) - D_0(p)\|
\end{equation} 

\item SSIM \cite{wang2004image}:
\begin{equation}
    \frac{(2\mu_{D_0(p)} \mu_{D(p)} + c_1)(2 \sigma_{D_0(p)D(p)} + c_2)}{(\mu_{D_0(p)}^2 + \mu_{D(p)}^2 + c_1)(\sigma_{D_0(p)}^2 + \sigma_{D(p)}^2+c_2)}
\end{equation}
where $c_1 = 0.0001, c_2 = 0.0009$.

\item $\delta_t$: percentage of pixels within the error range $t$, the error range is defined by equation \ref{eq:delta}.
\begin{equation}\label{eq:delta}
    max(\frac{D(p)}{D_0(p)}, \frac{D_0(p)}{D(p)}) < t
\end{equation} 
where $t \in \{1.05, 1.10, 1.25, 1.25^2, 1.25^3\}, p \in obs$.

\end{itemize}

\begin{table*}[t]
    \centering
    \begin{tabular}{|l|cc|cccccc|}\hline
         Model&RMSE$\downarrow$&Mean$\downarrow$&SSIM$\uparrow$&$1.05\uparrow$&$1.10\uparrow$&$1.25\uparrow$&$1.25^2\uparrow$&$1.25^3\uparrow$  \\ \hline
         Bilateral&1.978&0.774&0.507&0.385&0.497&0.613&0.689&0.730\\ 
         MRF \cite{harrison2010image}&1.675&0.618&0.692&0.506&0.556&0.651&0.780&0.856\\ 
         AD \cite{liu2013guided} & 1.653 & 0.610 & 0.696 & 0.503 & 0.560 & 0.663 & 0.792 & 0.861 \\
         
         FCN&1.262&0.517&0.605&0.397&0.527&0.681&0.808&0.868\\ 
         Zhang \cite{zhang2018deep} \footnotemark&1.316&0.461&0.762&0.657&0.708&0.781&0.851&0.888\\ 
         {\bf Ours}&{\bf 1.092}&{\bf 0.342}&{\bf 0.799}&{\bf 0.661}&{\bf 0.750}&{\bf 0.850}&{\bf 0.911}&{\bf 0.936}\\ \hline
    \end{tabular}
    \caption{We achieve state-of-the-art performance on Matterport3D dataset. We compare our method with some inpainting methods, including traditional methods, such as anisotropic diffusion, joint bilateral filter, FCN-based methods (ResNet18) and Zhang \cite{zhang2018deep} et al. The result shows our significant improvement on all of the evaluation metrics.}
    \label{tab:performance}
\end{table*}
\footnotetext{We reproduce Zhang's work on github and find they use root median square error instead of root mean square error. Thus, we show the results reproduced ourselves.}

\subsection{Performance} \label{tag:Performance}
As shown in Table \ref{tab:performance}, our proposed method defeats all other works, including traditional methods and previous state-of-the-art \cite{zhang2018deep}, and shows significant performance gain on RMSE and Mean error. Not only SSIM metric but also most delta values show improvement with boundary consistency and self-attention. SSIM is an index standing for the depth quality and structure. Delta percentage is a measurement of relative depth difference, which means the pixels closer to depth sensor requires more precise prediction. Our model attains the highest performance on all of the above evaluation metrics.

\begin{figure}[h!]
    \centering
    \includegraphics[width=\linewidth]{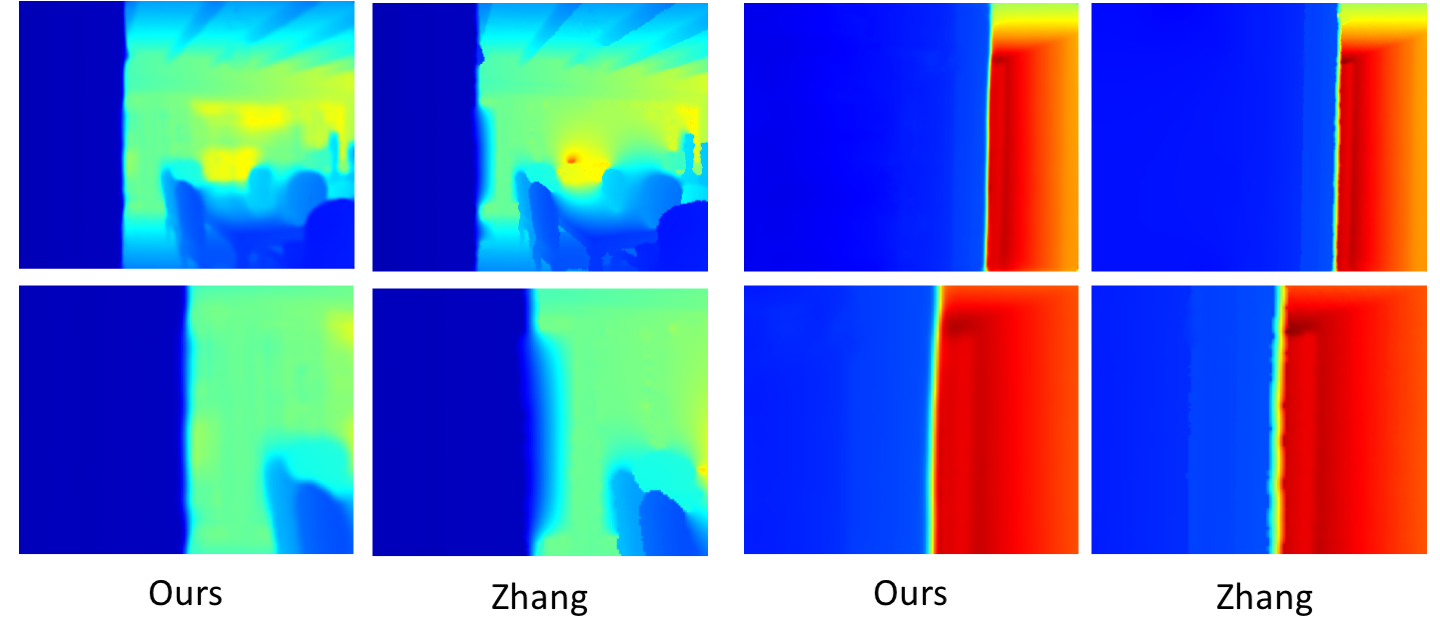}
    \caption{With the help of boundary consistency and SSIM loss, our method produces depth map of clearer structures. The upper row is original output depth, and the lower row is the partial amplification of the upper one. In the left two columns, our wall structure is clear. In the right two columns, our depth quality is higher, since boundary detail is not blurred.}
    \label{fig:performance_detail}
\end{figure}

Figure \ref{fig:performance} shows the visualization results. Previous  methods sometimes misunderstands the geometric meaning of one scene once raw depth is missing large coverage of depth value, which cause copy and interpolation. On the contrary, our self-attention network realizes the geometric features from RGB, normal and boundary and produces depth maps with more accurate geometric meaning than others. The red circle in Figure \ref{fig:performance} shows the power of self-attention mechanism on our task. Our network notices detailed geometric properties, like walls and corridors. Besides, with boundary consistency Figure \ref{fig:performance_detail}, our network learns architectural details, like the wall's edge, and produces depth with desirable quality.

\subsection{Ablation Studies}
For better understanding of our works, we investigate the effects of each component of our method. The following ablation study concentrates on the following three things: self-attention, SSIM loss and boundary consistency.

\subsubsection{Self-Attention}
The first two rows of Table \ref{tab:ablation} exam whether self-attention mechanism has positive effect on the result of complete depth. Indeed, self-attention makes huge gains of performance over traditional FCN models. In this setting, we take ResNet18 \cite{he2016deep}, which has similar parameters, as the classic FCN method. 

As we can see in Figure \ref{fig:structral}, on the right hand side of corridors next to the pillar, model with self-attention(SA) notices those depth value should be distant, while without SA misunderstands the geometric meaning. We believe that the improvement comes from the attention on convolution features to helps the model focus on important regions and features. In this case, self-attention mechanism helps model to learn more geometric meanings. 

\subsubsection{SSIM Loss}
By adding small weight of SSIM loss to optimize, the self-attention network learns to balance structural information without degradation much on RMSE and delta percentage. The straightforward results that SSIM score improves $8.6\%$ after adding SSIM loss represent network successfully produce higher quality of depth map values. 

As we can see in Figure \ref{fig:structral}, on the second row of columns w/SA and SA+SSIM, the background surface is smooth and accurate. Also, the edge detected from Sobel shows the enhancement of depth image quality and reduction in noise, which can be seen from the layered lines on the floor and ceiling.

\subsubsection{Boundary Consistency}
The last two rows of Table \ref{tab:ablation} show boundary consistency loss helps a lot in generating better depth map. There is extraordinary improvement in SSIM and delta percentage. $65.7\%$ of $\delta 1.05$ and $75.8\%$ of $\delta 1.10$ require highly sophisticated prediction of depth map for most of depth values. Besides, $3\%$ of SSIM gains implies boundary consistency helps enhance the structure and sharpness of boundary and validates the effectiveness of boundary consistency. 

\begin{figure*}[ht!]
    \centering
    \includegraphics[width=\textwidth]{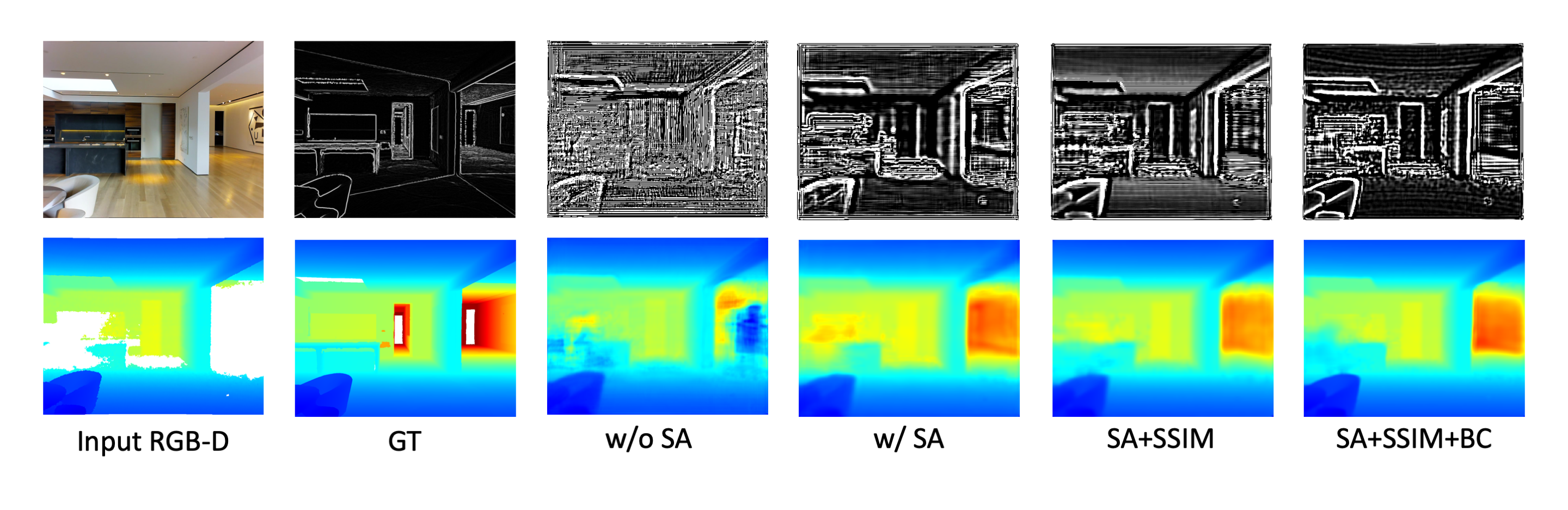}
    \caption{In the visualization result, we can see our proposed SA+SSIM+BC (self-attention, SSIM loss and boundary consistency) performs the best. The upper row is the Sobel edge detection results from the depth images of bottom row. From left to right images, we can see the depth background surface and bottom left sofa gradually improve both in structure and boundary. Model can learn better geometric meanings and semantics with the help of SA, SSIM and BC.}
    \label{fig:structral} 
\end{figure*}

As we can see in Figure \ref{fig:structral}, the structure of the bottom left sofa is clearer both in depth map and Sobel edge detection. Besides, the floor and ceiling are smoother with boundary consistency shown in the Sobel edge detection.

\begin{table*}[h!]
    \centering
    \begin{tabular}{|l|cc|cccccc|}\hline
         Model&RMSE$\downarrow$&Mean$\downarrow$&SSIM$\uparrow$&$1.05\uparrow$&$1.10\uparrow$&$1.25\uparrow$&$1.25^2\uparrow$&$1.25^3\uparrow$  \\ \hline
         W/O SA&1.262&0.517&0.605&0.397&0.527&0.681&0.808&0.868\\ 
         SA&1.095&0.400&0.706&0.497&0.629&0.785&0.881&0.923\\ 
         SA+SSIM&1.096&0.397&0.767&0.488&0.626&0.787&0.884&0.926\\ 
         {\bf SA+SSIM+BC}&{\bf 1.092}&{\bf 0.342}&{\bf 0.799}&{\bf 0.661}&{\bf 0.750}&{\bf 0.850}&{\bf 0.911}&{\bf 0.936}\\ \hline
    \end{tabular}
    \caption{Effect of each component of our model. We show that our proposed self-attention mechanism (SA), SSIM loss and boundary consistency (BC) enhances completion performance.}
    \label{tab:ablation}
\end{table*}

\subsection{Comparison}
In this section, we compare our main advantages over previous state-of-the-art work \cite{zhang2018deep}. Mentioned in Section \ref{tag:Performance}, self-attention mechanism can be easier to learn geometric information than other methods. We find Zhang's method depended heavily on raw depth. Once there are large holes in raw depth, like views of a wide and deep space, Zhang's method fell short of understanding the geometric meaning and easily fit onto noises in raw depth. The results seems to be simple interpolation but not realization of the whole view via clues from RGB and depth representation features.

Running real-time of sensing depth and training time can be an issue for most depth completion models. Our work is end-to-end trainable and training time is faster than optimization-based algorithms. Zhang proposed model that was not end-to-end trainable and used global optimization. Our proposed method overcomes those difficulties and outperforms previous works.

\subsection{Attention Map}
\begin{figure}[h]
    \centering
    \includegraphics[width=\columnwidth]{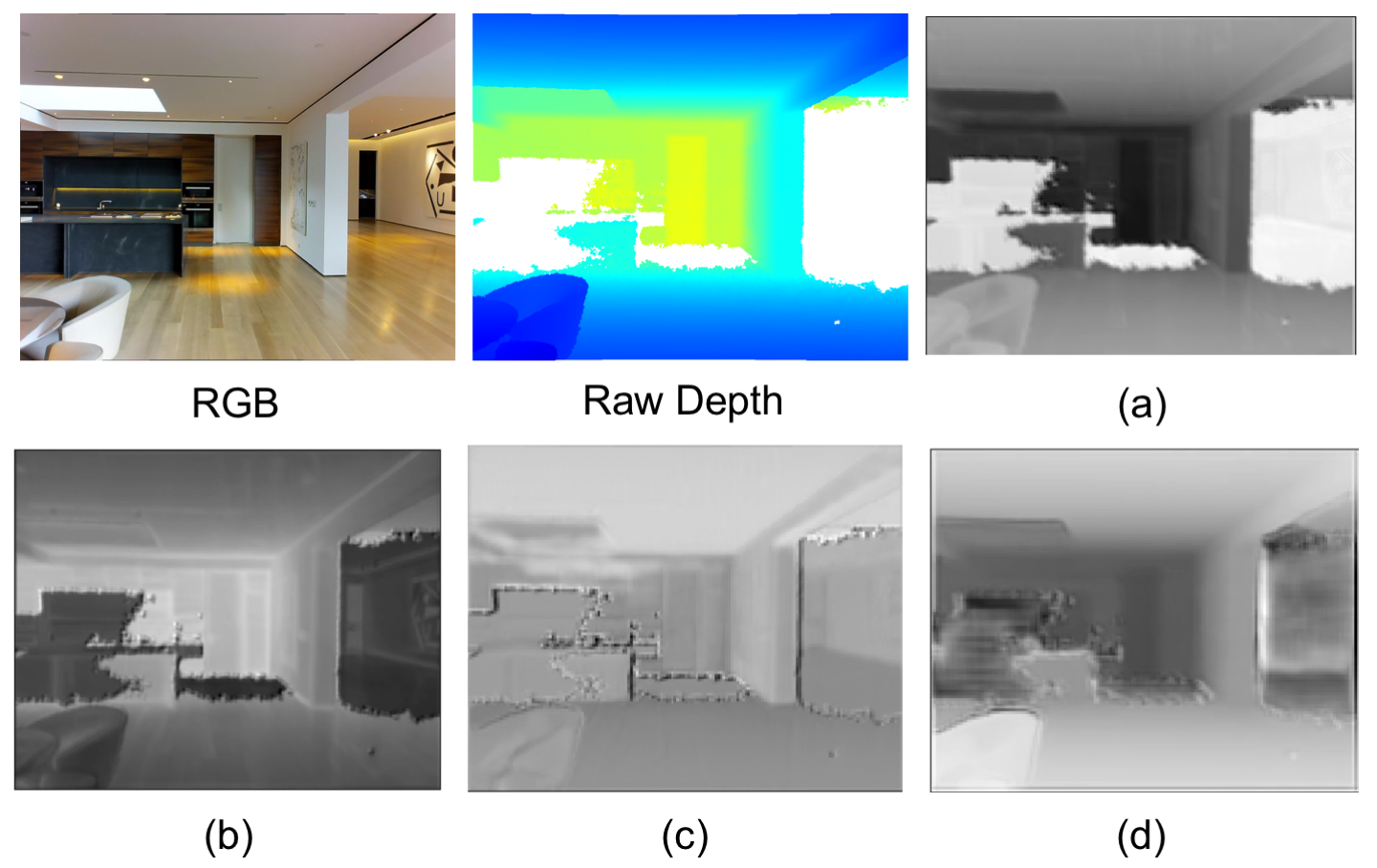}
    \caption{Visualization of some self-attention maps. Some filters pays attention on the missing holes (a), and some on the background and structures (b). More importantly, some filters pay attention on the semantics and object boundaries (c). For decoding phase, (d) shows the geometric meaning that the missing depth value behind the white walls at the right of the image should be relatively distant.}
    \label{fig:attention}
\end{figure}

To show the self-attention mechanism can capture the structure information, we randomly sample the attention output from our network. The result is shown in Figure \ref{fig:attention}. The attention map is able to distinguish the occlusion boundaries and the geometric shape of the objects in the scene.

\section{Conclusion}
In this work, we propose to use self-attention mechanism and boundary consistency to improve performance in depth completion task. Self-attention mechanism allows network to learn more geometric meanings and complete depth with more precise values. Boundary consistency improves both the depth boundary and image quality, producing clearer and sharper structures. Extensive experiments demonstrate that our proposed method reach state-of-the-art on Matterport3D dataset. Moreover, ablation study validates the effectiveness of each component we propose. 

\section*{Acknowledgement}
This work was supported in part by the Ministry of Science and Technology, Taiwan, under Grant MOST 108-2634-F-002-004, FIH Mobile Limited, and Qualcomm Technologies, Inc., under Grant NAT-410477. We are grateful to the National Center for High-performance Computing.

{\small
\bibliographystyle{ieee}
\bibliography{egbib}
}

\end{document}